\documentclass[12pt]{iopart}
\usepackage{amssymb}
\usepackage{bm}
\usepackage{siunitx}
\usepackage[hyperfootnotes=false]{hyperref}
\usepackage[usenames,dvipsnames]{xcolor}
\usepackage{xspace}
\usepackage{tikz}
\usepackage{subfigure}
\usepackage{titlesec}
\usepackage{soul}
\usepackage{booktabs}    
\usepackage{mleftright}
\usepackage{fontawesome5}
\usepackage{cite}

\expandafter\let\csname equation*\endcsname\relax
\expandafter\let\csname endequation*\endcsname\relax

\usepackage{amsmath}
\usepackage[capitalise]{cleveref}

\hypersetup{
    colorlinks=true,       
    linkcolor=Mahogany,          
    citecolor=Mahogany,        
    urlcolor=Mahogany           
}

\newcommand{\MYhref}[3][blue]{\href{#2}{\color{#1}{#3}}}%

\newcommand\funop[1]{\mathop{{}#1}}

\newcommand{\mrm}[1]{\mathrm{#1}}
\newcommand{\nsi}[1]
{\varepsilon_{#1}\xspace}
\newcommand{\bmath}[1]{\boldsymbol{#1}}

\newcommand{\usim}{\mathord{\sim}}

\definecolor{orcid-green}{RGB} {166, 206, 57}

\crefname{appendix}{}{}

\begin{document}

\title[Fast Inference for Neutrino NSI]{Fast Bayesian Inference for Neutrino Non-Standard Interactions at Dark Matter Direct Detection Experiments}

\author{Dorian W.~P.~Amaral$^{1}$\footnotemark \MYhref[orcid-green]{https://orcid.org/0000-0002-1414-932X}{\faOrcid}, Shixiao Liang$^{1}$\footnotemark[\value{footnote}]\MYhref[orcid-green]{https://orcid.org/0000-0003-0116-654X}{\faOrcid},  Juehang Qin$^{1}$\footnotemark[\value{footnote}] \MYhref[orcid-green]{https://orcid.org/0000-0001-8228-8949}{\faOrcid},\\ and Christopher Tunnell$^{1,2}$ \MYhref[orcid-green]{https://orcid.org/0000-0001-8158-7795}{\faOrcid}}

\address{$^1$ Department of Physics and Astronomy,  \\ Rice University,
Houston, TX, 77005, U.S.A.}
\address{$^2$Department of Computer Science,\\
Rice University,
Houston, TX, 77005, U.S.A.}

\eads{\mailto{dorian.amaral@rice.edu}, \mailto{liangsx@rice.edu}, \\ \mailto{qinjuehang@rice.edu}, \mailto{tunnell@rice.edu}}

\footnotetext{Equal contribution.}

\begin{abstract}
Multi-dimensional parameter spaces are commonly encountered in physics theories that go beyond the Standard Model. However, they often possess complicated posterior geometries that are expensive to traverse using techniques traditional to astroparticle physics. Several recent innovations, which are only beginning to make their way into this field, have made navigating such complex posteriors possible. These include GPU acceleration, automatic differentiation, and neural-network-guided reparameterization. We apply these advancements to dark matter direct detection experiments in the context of non-standard neutrino interactions and benchmark their performances against traditional nested sampling techniques when conducting Bayesian inference. Compared to nested sampling alone, we find that these techniques increase performance for both nested sampling and Hamiltonian Monte Carlo, accelerating inference by factors of $\usim 100$ and $\usim 60$, respectively. As nested sampling also evaluates the Bayesian evidence, these advancements can be exploited to improve model comparison performance while retaining compatibility with existing implementations that are widely used in the natural sciences. Using these techniques, we perform the first scan in the neutrino non-standard interactions parameter space for direct detection experiments whereby all parameters are allowed to vary simultaneously. We expect that these advancements are broadly applicable to other areas of astroparticle physics featuring multi-dimensional parameter spaces.
\end{abstract}

\section{Introduction}
\label{sec:intro}
The Standard Model (SM) of particle physics, our best description of the fundamental building blocks of the Universe, is an incomplete description of nature. To better capture reality, theories that go beyond the Standard Model (BSM) are needed~\cite{Murayama:2007ek,Allanach:2016yth,Lee:2019zbu}. These theories often feature multi-dimensional parameter spaces that must be explored to determine regions that are allowed by the observed data~\cite{Fan:2010gt,Feroz:2011bj,Levi:2018nxp,Arguelles:2022lzs,Isidori:2023pyp,Amaral:2023tbs}. However, methods for doing so that are traditional to the astroparticle physics community, such as nested sampling (NS), are often inefficient at navigating complicated posterior geometries \cite{2018arXiv180306387C,Cerdeno:2018bty,2020arXiv201215286A, Trotta2008TheIO}. This problem is particularly acute in the limit-setting regime, where no significant novel signal is observed and the data is used to restrict the domain of the theory. In this regime, the likelihood is often diffuse and lacks strong modes, rendering it difficult to explore. The problem is exacerbated when the likelihood is expensive to evaluate, as is often true due to the complexity of theoretical frameworks. These issues make it challenging to diagnose convergence failures, incorporate new experimental data, and produce timely studies.

These problems can be tackled using a combination of methods. Leveraging recent powerful numerical computing frameworks, such as \texttt{JAX}~\cite{jax2018github}, and probabilistic programming frameworks, such as \texttt{numpyro}~\cite{bingham2018pyro, phan2019composable}, allows us to automatically differentiate likelihood functions and thus sample efficiently using Hamiltonian Monte Carlo (HMC)~\cite{Duane:1987de, Neal2011MCMCUH} and the No-U-Turn Sampler (NUTS)~\cite{hoffman2014no}. For likelihood functions that are computationally expensive but vectorizable, GPU acceleration can also confer a significant performance boost. Finally, normalizing flows can be used as a reversible transformation to obtain a posterior distribution that is easier to explore in a technique known as neural transport (NeuTra)~\cite{2019arXiv190303704H}. 

The framework of neutrino non-standard interactions (NSI) provides an excellent physics playground on which to benchmark these methods. This framework is a low-energy effective field theory description of BSM neutrino physics, encapsulating the novel interactions that neutrinos could undergo with SM fermions~\cite{Wolfenstein:1977ue,Guzzo:1991hi,Guzzo:1991cp,Gonzalez-Garcia:1998ryc,Guzzo:2000kx,Guzzo:2001mi,Gonzalez-Garcia:2004pka}. It features at least $8$ independent parameters: $6$ Wilson coefficients and $2$ angles describing the strength of the new physics interaction with either nucleons or electrons~\cite{Amaral:2023tbs,Coloma:2023ixt}. To explore the NSI landscape fully, we consider dark matter direct detection experiments, which are sensitive to both nuclear recoils (NRs) and electron recoils (ERs) from neutrinos.

While predominantly employed in the search for dark matter, direct detection experiments have improved their sensitivity to the point that they are beginning to approach the `neutrino fog'. This marks a region of the WIMP (Weakly Interacting Massive Particle) dark matter parameter space where the rate of NRs due to neutrinos becomes non-negligible, leading to a cumbersome background for dark matter searches~\cite{OHare:2021utq}. However, this background can be re-imagined as an invaluable signal for direct detection experiments, providing them with the opportunity to probe both SM and BSM neutrino physics. Indeed, the first indications of solar neutrino interactions have recently been observed in direct detection experiments based on dual-phase xenon time projection chambers (TPCs)~\cite{PandaX:2024muv,XENON:2024ijk}. ER events are also possible with solar neutrinos and have a different signature from NRs; they can be measured independently, and serve as complementary probes of the neutrino NSI parameter space. These interactions between solar neutrinos and the target material can be used to explore the nature of novel neutrino physics in the NSI framework~\cite{Amaral:2023tbs,Dutta:2020che,Dutta:2017nht}.

In this work, we study the use of GPU acceleration, differentiable likelihoods, and NeuTra to constrain neutrino non-standard interactions using results from dark matter direct detection experiments. We begin by studying the performance boosts that can be achieved using differentiable likelihoods and NeuTra with a synthetic problem based on a multidimensional Gaussian fit. We then apply these methods and GPU acceleration within the context of the higher dimensional NSI framework. We perform statistical inference within this framework using results from the astroparticle experiments XENON1T~\cite{XENON:2020gfr} and PandaX-4T~\cite{Lu:2024ilt} to obtain the first NSI parameter space scan for direct detection experiments whereby all the parameters are allowed to vary simultaneously. We provide our code in \href{https://github.com/RiceAstroparticleLab/ML-NSI}{\faGithub}.

\section{Sampling Methods}
\label{sec:background}

\subsection{Nested sampling}
\label{subsubsec:ns}
 
Nested sampling (NS) is a family of Monte Carlo methods for calculating integrals over a parameter space~\cite{skilling2004, Buchner2023}. It can also be used for Bayesian computation and has gained popularity for its success in multi-modal problems and its self-tuning feature~\cite{skilling2006}. NS has many applications in various fields, such as biology~\cite{russel2019model, arvindekar2024optimizing}, linguistics~\cite{sagart2019dated, robbeets2018bayesian} and the physical sciences~\cite{ashton2022nested}---including particle physics~\cite{Yallup:2022yxe, Dutta:2020che}. 

Consider a parameter space $\boldsymbol{\theta}$ for which we wish to compute the posterior distribution $p(\boldsymbol{\theta} | \boldsymbol{X})$, where $\boldsymbol{X}$ represents the experimental data. NS works by iteratively sampling live points from the prior distribution, $\pi(\boldsymbol{\theta})$, commencing by sampling $N$ live points within its support. Following this, the live point with the lowest likelihood value, $L_1$, is removed, and a new live point is sampled from the restricted parameter space, where the likelihood for the parameters in this space must be higher. This process continues, with sampling taking place within the contour defined by the live point with the lowest likelihood in the $i^{\text{th}}$ iteration, $L_i$. Concretely, the distribution being sampled from in the $i^{\text{th}}$ iteration is
\begin{equation}
    p_i(\boldsymbol{\theta}) \propto \begin{cases}
        \pi(\boldsymbol{\theta}) \quad &\text{if}\quad p(\boldsymbol{X} | \boldsymbol{\theta}) > L_i \\
        0 \quad &\text{if}\quad p(\boldsymbol{X} | \boldsymbol{\theta}) \leq L_i
    \end{cases}\,,
\end{equation}
where $p(\boldsymbol{X} | \boldsymbol{\theta})$ is the likelihood of the data $\bmath{X}$ given the parameters $\bmath{\theta}$. Increasing the likelihood threshold reduces the volume of the likelihood-restricted parameter space by a factor that obeys a $\text{Beta}(1, N)$ distribution. This factor is approximately given by $\Delta V_i = V_i/N$, where $V_i$ is the volume of the likelihood-restricted parameter space in the $i^{\text{th}}$ iteration.

Nested sampling stops when a predetermined halting criterion is met. One example of such a criterion is to require that the removed points represent a certain fraction of the total Bayesian evidence integral. The Bayesian evidence, $\mathcal{Z}$, can be estimated using the removed points and the remaining live points at termination via 
\begin{equation}
    \mathcal{Z} \equiv \int p(\boldsymbol{\theta} | \boldsymbol{X})\,
    \mathrm{d}\boldsymbol{\theta} \simeq \sum_{i=1}^{N_{\text{iter}}} \Delta V_i  L_i + \frac{V_{\text{end}}}{N}\sum_{j=1}^N L_j,
\end{equation}
where the sum over $j$ represents a sum over the likelihoods of the remaining live points at termination, $N_{\text{iter}}$ is the total number of iterations, and $V_{\text{end}}$ is the remaining volume at termination.
The posterior distribution can then be inferred using the removed points and the remaining live points at termination weighted by $\Delta V_i  L_i$ and ${V_{\text{end}}}/{N}$, respectively. The ability to provide both the evidence and the posterior distribution makes NS a powerful method for Bayesian computation.

\subsection{HMC/NUTS}
\label{subsubsec:hmc}

Hamiltonian Monte Carlo is a powerful sampling algorithm that leverages gradient information to efficiently explore complex probability distributions~\cite{Duane:1987de, Neal2011MCMCUH, Betancourt:2017ebh}. HMC offers improved performance compared to traditional Markov chain Monte Carlo (MCMC) methods using a simulation of Hamiltonian dynamics to explore a target parameter space more efficiently than random walks. However, The effectiveness of HMC relies on careful tuning. The NUTS sampler with dual-averaging alleviates this need by adaptively adjusting HMC parameters~\cite{pmlr-v130-hoffman21a, pmlr-v151-hoffman22a}, and it has been widely adopted and implemented across various probabilistic programming frameworks~\cite{hoffman2014no, bingham2018pyro, phan2019composable, carpenter2017stan, pymc2023}.

HMC is extensively used across both the natural and social sciences, and it has been recently employed for analyzing neutrino experiment results~\cite{Duane:1987de, Freedman:2021ahq, sharma2021understanding, Wang2022BrainPyAF, Fazio2024GaussianDS, NOvA:2023iam}. Because HMC is a gradient-based sampling method, it requires the model to either support automatic differentiation or have known derivatives. Interest in it has thus led to new implementations of models using frameworks compatible with automatic differentiation~\cite{Campagne:2023ter, Ruiz-Zapatero:2023hdf}.

\subsection{Neural Transport}
\label{subsubsec:neutra}

While NS and HMC are better at exploring complex posteriors than simpler methods, such as basic rejection sampling with uniform proposals, sampling efficiency still drops for sufficiently challenging posteriors in multiple dimensions~\cite{Buchner2023, Papaspiliopoulos2007AGF}. Under these circumstances, reparameterization can improve the performance of inference algorithms~\cite{pmlr-v119-gorinova20a}, which can be non-trivial for likelihoods that cannot be easily expressed as hierarchical Bayesian models.

One way to remedy this is to use a reversible transformation to map a posterior distribution onto one that is easier to sample, such as a standard normal distribution~\cite{2019arXiv190303704H}. Instead of directly sampling from the posterior $p(\boldsymbol{\theta} | \boldsymbol{X})$ using HMC or NS, we can consider a bijective function $\boldsymbol{f}$ that defines the reparameterization of our distribution. This function acts on the reparameterized latent space $\boldsymbol{z}$, such that $\boldsymbol{\theta}=\boldsymbol{f}(\boldsymbol{z})$. We can then instead sample from the distribution
\begin{equation}
\label{eq:latent-transform}
    p(\boldsymbol{z}) \equiv p(\boldsymbol{\theta} = \boldsymbol{f}(\boldsymbol{z}) | \boldsymbol{X})\left|\frac{\partial \boldsymbol{f}}{\partial \boldsymbol{z}}\right|\,,
\end{equation}
where $|\partial \boldsymbol{f} / \partial \boldsymbol{z}|$ is the determinant of the Jacobian of the parameter transformation. \cref{eq:latent-transform} generally follows from the reparameterization of probability distributions, as shown in \cref{appendix:reparam}.

Subsequently, the pushforward $\bmath{f}(\boldsymbol{z})$ can be used to obtain the desired samples of $p(\boldsymbol{\theta} | \boldsymbol{X})$. Normalizing flows are a good candidate to parameterize the bijective function, as they can be highly expressive and can be fit by minimizing the Kullback–Leibler (KL) divergence. Using them to reparameterize Bayesian models is known as Neural Transport (NeuTra), as introduced in~\cite{2019arXiv190303704H}. There, NeuTra with inverse autoregressive flows was shown to improve sampling performance in problems with difficult posterior distributions. Instead of inverse autoregressive flows, we use block neural autoregressive flows, which demonstrate improved performance over earlier methods~\cite{2019arXiv190404676D}.

To train a normalizing flow that can transform the posterior distribution of interest into a standard normal distribution, we maximize the evidence lower bound (ELBO)~\cite{2019arXiv190303704H, Ranganath2013BlackBV}, which is equivalent to minimizing the KL divergence~\cite{kingma2013auto}. Given $\boldsymbol{f}_\phi (\boldsymbol{z})$---the transformation represented by a normalizing flow with parameters $\phi$---the distribution over $\boldsymbol{\theta}$ is 
\begin{equation}
    q_\phi(\boldsymbol{\theta}) \equiv q_\phi(\boldsymbol{z})\mleft|\frac{\partial \boldsymbol{f}}{\partial \boldsymbol{z}}\mright|^{-1},
\end{equation}
where $q_\phi(\boldsymbol{z})$ is the target distribution in the transformed space, which is a standard normal distribution in our case. The ELBO can then be written as
\begin{equation}\label{eq:elbo}
    \mathcal{L}(\phi) = \mathbb{E}_{q_\phi(\boldsymbol{z})}\mleft[\log\mleft(p(\boldsymbol{X}, \boldsymbol{f}_\phi (\boldsymbol{z}))\mright) - \log\mleft(q_\phi(\boldsymbol{z})\left|\frac{\partial \boldsymbol{f}}{\partial \boldsymbol{z}}\right|^{-1}\mright)\mright]\,,
\end{equation}
and we can thus estimate it using a set of samples from a standard normal distribution.

\section{Neutrino Non-Standard Interactions and Direct Detection Experiments}
\label{subsec:nsi}

\subsection{Neutrino Non-Standard Interactions (NSI)} 

\begin{figure}[t!]
    \begin{center}
    \begin{minipage}{0.5\textwidth}
    \input{figs/sun.tikz}
    \end{minipage}
    \hspace{1.05cm}
    \begin{minipage}{0.375\textwidth}
    \tikzset{every picture/.style={line width=0.75pt}} 

\begin{tikzpicture}[x=0.75pt,y=0.75pt,yscale=-0.7,xscale=0.7]

\draw  [color={rgb, 255:red, 0; green, 0; blue, 0 }  ,draw opacity=1 ][fill={rgb, 255:red, 128; green, 128; blue, 128 }  ,fill opacity=0 ] (229.06,188.86) .. controls (229.06,126.83) and (279.34,76.55) .. (341.36,76.55) .. controls (403.39,76.55) and (453.67,126.83) .. (453.67,188.86) .. controls (453.67,250.88) and (403.39,301.16) .. (341.36,301.16) .. controls (279.34,301.16) and (229.06,250.88) .. (229.06,188.86) -- cycle ;
\draw  [draw opacity=0][fill={rgb, 255:red, 125; green, 156; blue, 186 }  ,fill opacity=0.5 ][dash pattern={on 4.5pt off 4.5pt}] (435.59,174.38) .. controls (436.22,179.48) and (436.54,184.7) .. (436.54,190) .. controls (436.54,250.93) and (393.59,300.36) .. (340.5,300.67) -- (340,190) -- cycle ; \draw  [color={rgb, 255:red, 0; green, 0; blue, 0 }  ,draw opacity=1 ][dash pattern={on 4.5pt off 4.5pt}] (435.59,174.38) .. controls (436.22,179.48) and (436.54,184.7) .. (436.54,190) .. controls (436.54,250.93) and (393.59,300.36) .. (340.5,300.67) ;  
\draw    (340,320) -- (340,74.5) -- (340,62) ;
\draw [shift={(340,60)}, rotate = 90] [color={rgb, 255:red, 0; green, 0; blue, 0 }  ][line width=0.75]    (7.65,-2.3) .. controls (4.86,-0.97) and (2.31,-0.21) .. (0,0) .. controls (2.31,0.21) and (4.86,0.98) .. (7.65,2.3)   ;
\draw    (210,190) -- (468,190) ;
\draw [shift={(470,190)}, rotate = 180] [color={rgb, 255:red, 0; green, 0; blue, 0 }  ][line width=0.75]    (7.65,-2.3) .. controls (4.86,-0.97) and (2.31,-0.21) .. (0,0) .. controls (2.31,0.21) and (4.86,0.98) .. (7.65,2.3)   ;
\draw    (229,233.5) -- (453.64,146.22) ;
\draw [shift={(455.5,145.5)}, rotate = 158.77] [color={rgb, 255:red, 0; green, 0; blue, 0 }  ][line width=0.75]    (7.65,-2.3) .. controls (4.86,-0.97) and (2.31,-0.21) .. (0,0) .. controls (2.31,0.21) and (4.86,0.98) .. (7.65,2.3)   ;
\draw  [draw opacity=0] (452.45,189.56) .. controls (452.48,189.83) and (452.49,190.09) .. (452.49,190.36) .. controls (452.49,206.93) and (403.22,220.36) .. (342.45,220.36) .. controls (317.86,220.36) and (295.16,218.17) .. (276.84,214.45) -- (342.45,190.36) -- cycle ; \draw  [color={rgb, 255:red, 0; green, 0; blue, 0 }  ,draw opacity=1 ] (452.45,189.56) .. controls (452.48,189.83) and (452.49,190.09) .. (452.49,190.36) .. controls (452.49,206.93) and (403.22,220.36) .. (342.45,220.36) .. controls (317.86,220.36) and (295.16,218.17) .. (276.84,214.45) ;  
\draw  [draw opacity=0][dash pattern={on 4.5pt off 4.5pt}] (405.85,165) .. controls (434.05,170.43) and (452.45,179.38) .. (452.45,189.5) .. controls (452.45,189.52) and (452.45,189.54) .. (452.45,189.56) -- (342.25,189.5) -- cycle ; \draw  [color={rgb, 255:red, 0; green, 0; blue, 0 }  ,draw opacity=1 ][dash pattern={on 4.5pt off 4.5pt}] (405.85,165) .. controls (434.05,170.43) and (452.45,179.38) .. (452.45,189.5) .. controls (452.45,189.52) and (452.45,189.54) .. (452.45,189.56) ;  
\draw  [draw opacity=0][fill={rgb, 255:red, 145; green, 213; blue, 216 }  ,fill opacity=0.5 ] (402.33,164.99) .. controls (433.03,170.37) and (453.25,179.55) .. (453.25,189.97) .. controls (453.25,206.54) and (402.11,219.97) .. (339.03,219.97) .. controls (315.63,219.97) and (293.88,218.12) .. (275.77,214.95) -- (339.03,189.97) -- cycle ; \draw  [draw opacity=0] (402.33,164.99) .. controls (433.03,170.37) and (453.25,179.55) .. (453.25,189.97) .. controls (453.25,206.54) and (402.11,219.97) .. (339.03,219.97) .. controls (315.63,219.97) and (293.88,218.12) .. (275.77,214.95) ;  
\draw  [draw opacity=0][fill={rgb, 255:red, 125; green, 156; blue, 186 }  ,fill opacity=0.5 ][dash pattern={on 4.5pt off 4.5pt}] (340.18,76.36) .. controls (388.94,76.47) and (429.21,119.15) .. (435.62,174.53) -- (340,190) -- cycle ; \draw  [color={rgb, 255:red, 0; green, 0; blue, 0 }  ,draw opacity=1 ][dash pattern={on 4.5pt off 4.5pt}] (340.18,76.36) .. controls (388.94,76.47) and (429.21,119.15) .. (435.62,174.53) ;  
\draw [color={rgb, 255:red, 0; green, 0; blue, 0 }  ,draw opacity=1 ]   (357.5,166.75) .. controls (365,173.25) and (364.5,181.25) .. (364.5,186) ;
\draw [color={rgb, 255:red, 0; green, 0; blue, 0 }  ,draw opacity=1 ][fill={rgb, 255:red, 0; green, 0; blue, 0 }  ,fill opacity=1 ][line width=0.75]  [dash pattern={on 4.5pt off 4.5pt}]  (340,190) -- (398.8,111.6) ;
\draw [shift={(400,110)}, rotate = 126.87] [fill={rgb, 255:red, 0; green, 0; blue, 0 }  ,fill opacity=1 ][line width=0.08]  [draw opacity=0] (12,-3) -- (0,0) -- (12,3) -- cycle    ;
\draw [color={rgb, 255:red, 0; green, 0; blue, 0 }  ,draw opacity=1 ]   (390,181.5) .. controls (392.25,183.5) and (392.5,185.75) .. (392,190) ;
\draw [color={rgb, 255:red, 0; green, 0; blue, 0 }  ,draw opacity=1 ][fill={rgb, 255:red, 208; green, 2; blue, 27 }  ,fill opacity=1 ] [dash pattern={on 0.84pt off 2.51pt}]  (400,110) -- (400,180) ;
\draw [color={rgb, 255:red, 0; green, 0; blue, 0 }  ,draw opacity=1 ] [dash pattern={on 0.84pt off 2.51pt}]  (340,190) -- (400,180) ;
\draw  [draw opacity=0][fill={rgb, 255:red, 0; green, 0; blue, 0 }  ,fill opacity=0 ] (277.02,214.76) .. controls (248.02,209.36) and (228.98,200.27) .. (228.98,189.97) .. controls (228.98,189.6) and (229.01,189.23) .. (229.06,188.86) -- (339.03,189.97) -- cycle ; \draw  [color={rgb, 255:red, 0; green, 0; blue, 0 }  ,draw opacity=1 ] (277.02,214.76) .. controls (248.02,209.36) and (228.98,200.27) .. (228.98,189.97) .. controls (228.98,189.6) and (229.01,189.23) .. (229.06,188.86) ;  
\draw  [draw opacity=0][fill={rgb, 255:red, 255; green, 255; blue, 255 }  ,fill opacity=0.07 ][dash pattern={on 4.5pt off 4.5pt}] (228.99,189.7) .. controls (229.51,173.26) and (278.58,159.97) .. (339.03,159.97) .. controls (362.8,159.97) and (384.81,162.02) .. (402.8,165.52) -- (339.03,189.97) -- cycle ; \draw  [color={rgb, 255:red, 0; green, 0; blue, 0 }  ,draw opacity=1 ][dash pattern={on 4.5pt off 4.5pt}] (228.99,189.7) .. controls (229.51,173.26) and (278.58,159.97) .. (339.03,159.97) .. controls (362.8,159.97) and (384.81,162.02) .. (402.8,165.52) ;  

\draw (462,128) node [anchor=north west][inner sep=0.75pt]   [align=left] {$\displaystyle \varepsilon _{\alpha \beta }^{e}$};
\draw (476,178) node [anchor=north west][inner sep=0.75pt]   [align=left] {$\displaystyle \varepsilon _{\alpha \beta }^{p}$};
\draw (331,32) node [anchor=north west][inner sep=0.75pt]   [align=left] {$\displaystyle \varepsilon _{\alpha \beta }^{n}$};
\draw (344.24,147) node [anchor=north west][inner sep=0.75pt]  [color={rgb, 255:red, 0; green, 0; blue, 0 }  ,opacity=1 ,rotate=-307.95] [align=left] {$\displaystyle \varepsilon _{\alpha \beta }$};
\draw (366.08,163) node [anchor=north west][inner sep=0.75pt]  [font=\footnotesize,color={rgb, 255:red, 0; green, 0; blue, 0 }  ,opacity=1 ] [align=left] {$\displaystyle \eta $};
\draw (397,176.83) node [anchor=north west][inner sep=0.75pt]  [font=\footnotesize,color={rgb, 255:red, 0; green, 0; blue, 0 }  ,opacity=1 ] [align=left] {$\displaystyle \varphi $};
\end{tikzpicture}
    \end{minipage}
    \end{center}
    \caption{\textbf{Left:} Experimental idea. Solar neutrinos collide with the target atoms in a direct detection experiment, causing either the nucleus or the electrons of the atom to recoil. \textbf{Right:} The neutrino non-standard interactions parameterization, which can be used to model new physics phenomena between neutrinos, neutrons ($n$), protons ($p$), and electrons ($e$) \cite{Amaral:2023tbs}. It is characterized by an overall interaction strength coordinate, $\nsi{\alpha\beta}$, and two angles describing how strong the interaction is with the neutron ($\eta$) and proton vs.~electron ($\varphi$).}
    \label{fig:astro-nsi}
\end{figure}

Neutrino physics that goes beyond the Standard Model can be described using the framework of neutrino non-standard interactions~\cite{Wolfenstein:1977ue,Guzzo:1991hi,Guzzo:1991cp,Gonzalez-Garcia:1998ryc,Guzzo:2000kx,Guzzo:2001mi,Gonzalez-Garcia:2004pka}. In this framework, novel neutrino interactions are parametrized within an $8$-dimensional space defined by the coordinates $(\nsi{\alpha\beta}, \eta, \varphi)$~\cite{Amaral:2023tbs}. Here, the $6$ parameters $\nsi{\alpha\beta}$ characterize the overall strength and type of interaction. There is one such parameter for each flavor combination $\alpha\beta$ ($ee$, $e\mu$, $e\tau$, etc.), with equal flavor indices meaning that the neutrino type remains the same after the interaction and unequal indices indicating that it changes. The remaining two parameters $(\eta,\varphi)$ are angles describing how strongly the new interaction takes place with protons, neutrons, and electrons. The angle $\eta$ models how strongly the interaction is with the neutron, and the angle $\varphi$ characterizes the strength of the interaction between the electrically charged proton and electron. We visualize the NSI parametrization in \cref{fig:astro-nsi} (right)

Concretely, the interactions with these  particles can be written via the transformation equations
\begin{equation}
\varepsilon^p_{\alpha\beta} = \sqrt{5}\nsi{\alpha\beta} \cos\eta \cos\varphi\,, \qquad
\varepsilon^e_{\alpha\beta} = \sqrt{5}\nsi{\alpha\beta} \cos\eta \sin\varphi\,, \qquad
\varepsilon^n_{\alpha\beta} = \sqrt{5}\nsi{\alpha\beta} \sin\eta\,,
    \label{eq:nsi-sphere}
\end{equation}
where the factor of $\sqrt{5}$ is a convention in the field \cite{Esteban:2018ppq,Amaral:2023tbs}. Except for this factor, this system of equations can be seen to be the projection from the `spherical' NSI parameterization onto a `Cartesian' one.

\subsection{Dual-phase Xenon Direct Detection Experiments}

Originally leading the hunt for dark matter, dual-phase xenon TPCs are beginning to have sensitivity to solar neutrinos~\cite{Aalbers:2022dzr}.
Two such experiments are XENON1T~\cite{XENON:2017lvq} and PandaX-4T~\cite{PandaX:2018wtu}. In these dual-phase xenon TPCs, the energy of the recoiling target due to an interaction with a solar neutrino is converted into scintillation and ionization signals. Measuring both signals allows us to detect low-energy recoils and discriminate between nuclear and electronic events. We provide a schematic of a solar neutrino interaction with a direct detection experiment in \cref{fig:astro-nsi} (left). We can count the number of nuclear recoils and/or electron recoilsdetermine whether the signal is consistent with the SM prediction. The NR and ER events constrain the neutrino NSI parameter space in complementary ways, leading to tighter limits or, in the case of a detection, smaller error bars~\cite{Amaral:2023tbs,Dutta:2020che}.

Within the NSI framework, we calculate the number of expected NRs and ERs using the \texttt{SNuDD} Python package \cite{SNuDD-code,Amaral:2023tbs}.  Given a choice of NSI parameters, it computes the number of expected recoils due to solar neutrinos, which is described by the expression~\cite{Coloma:2022umy},
\begin{equation}\label{eq:dr_gen}
    \frac{\mathrm{d} R}{\mathrm{d} E_R}=  N_T \int_{E_\nu^\mathrm{min}} \,\frac{\mathrm{d}{\phi_\nu}}{\mathrm{d}E_\nu}\,  \mathrm{Tr}\left[\boldsymbol{\rho}\, \,\frac{\mathrm{d}{\boldsymbol{\zeta}}}{\mathrm{d}E_R}\right] \, \mathrm{d} E_\nu\, .
\end{equation}
Here, $N_T$ is the number of targets, $\phi_\nu$ is the neutrino flux at the source, $\boldsymbol\rho$ is the neutrino density matrix at the experiment, $\boldsymbol\zeta$ is a generalized scattering cross section, $E_\nu$ is the energy of the incident neutrinos, and $E_\nu^\mathrm{min}$ is the minimum $E_\nu$ required to produce a target recoil energy of $E_R$. 

Both the density matrix and the generalized scattering cross section depend on the full set of NSI parameters. The density matrix dictates the relative neutrino flavor fractions we expect to see on Earth, and it is thus associated with the matter effects stemming from the Mikheyev-Smirnov-Wolfenstein (MSW) effect. Neutrino NSI lead to additional interactions between neutrinos and fermions within the solar medium described by an extra term in the matter Hamiltonian that ultimately alters the density matrix. Similarly, additional potential interactions due to neutrino NSI affect the scattering cross section at the detection point. Full details of both of these effects may be found in Ref.~\cite{Amaral:2023tbs}.

\section{Evaluation of Sampling Methods}
\label{sec:exps}

\subsection{Gaussian Fit with Unknown Mean and Variance}
\label{subsec:toyexp}

Before applying our sampling methods to the NSI problem, we compare them using a simpler synthetic problem. This is based on a Gaussian fit with unknown mean and standard deviation with a centered parameterization. This is chosen to demonstrate the performance of the various methods using a model that is known to be difficult to sample without reparameterization~\cite{pmlr-v119-gorinova20a}. The likelihood has a similar structure to the well-known Neal's funnel~\cite{neal2003slice}; however, this was chosen as a more difficult problem, as Neal's funnel was efficiently handled by nested sampling as implemented in \texttt{ultranest}~\cite{2021JOSS....6.3001B}.
Our model is given by
\begin{equation}
    \mu_i \sim \mathcal{N}(0, 10)\,, \quad
    \mathcal{C}_i \sim \mathcal{N}(0, 10)\,, \quad
    x_{ij} \sim \mathcal{N}(\mu_i, e^{\mathcal{C}_i})\,,
\label{eq:toy_model}
\end{equation}
where $\mu_i$ are means, $\mathcal{C}_i$ are log variances, and $x_{ij}$ are the observed variables. For our experiments, we fit 3-dimensional data generated from a standard normal distribution, and the observed dataset has 2 data points, such that $i \in \{1,2,3\}$ and $j \in \{1,2\}$. 

We sample from this model using NS and NUTS, both with and without NeuTra. For NS, the number of live points is $1600$ without neural transport and $2400$ with neural transport, as this is found to produce $1.4\times10^4$ effective samples in both cases. We use a sigmoid to transform the parameter space into the unit cube for NS runs. The log-determinant of the transformation Jacobian is summed into the log-density to ensure that the posterior is unaltered by this procedure. For NUTS, each of the $4$ chains we use has a length of 3000 warmup steps and 5000 samples with a target acceptance probability of 0.8.
We implement NeuTra with Block Neural Autoregressive Flows~\cite{2019arXiv190404676D} using two flows with hidden dimensions $[4, 4]$. We train the NeuTra model on the ELBO estimated using $30$ points for $5000$ epochs. We run all experiments on an Intel Xeon Gold 6230 CPU. The ground truth evidence and contour integrals are analytically evaluated where possible, and we use \texttt{torchquad}~\cite{Gomez_torchquad_Numerical_Integration_2021} to compute the remaining integrals via non-stochastic numerical integration; we describe this procedure in greater detail in~\cref{appendix:bayesian}.

\subsubsection{Results}

\begin{figure}
    \centering
\includegraphics{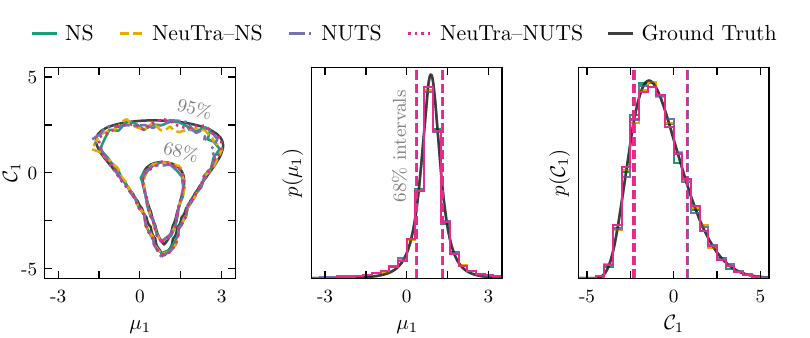}
    \caption{Marginalized representations of the inferred posteriors for our synthetic model, \cref{eq:toy_model}, using nested sampling (NS) and the No U-Turn Sampler (NUTS), both with and without neural transport (NeuTra). \textbf{Left:} The $68\%$ and $95\%$ highest posterior density region contours. \textbf{Middle and Right:} Probability density of the samples with $68\%$ equal-tailed intervals indicated as dashed lines. Only $\mu_1$ and $\mathcal{C}_1$ are shown for clarity; the full corner plot is given in \cref{appendix:corner}. As desired, our results show a high degree of overlap and all closely match the ground truth, derived in \cref{appendix:bayesian}.}
    \label{fig:toy}
\end{figure}

\begin{table}[t!]
\renewcommand{\arraystretch}{1.5}
    \begin{tabular*}{\textwidth}{@{\extracolsep{\fill}}lccc}
    \toprule
    \textbf{Method}       & \textbf{Wall Time/ESS~[ms]}  & \textbf{Eval./ESS} & \textbf{Div./ESS [$\mathbf{\times 10^{-2}}$]}\\
    \midrule
    NS           & $16.0 \pm 0.1$ & $20.8 \pm 0.4$ & N/A\\
    NeuTra--NS    & $12.0 \pm 0.6$ & $10 \pm 1$ & N/A\\
\midrule[0.25pt]
    NUTS         & $4.8 \pm 0.3$ & $86 \pm 6$ & $3 \pm 2$\\
    NeuTra--NUTS  & $1.4 \pm 0.2$ & $9.6 \pm 0.5$ & $0$\\
    \bottomrule
  \end{tabular*}
  \vspace{1ex}
  
  \caption{Synthetic model performance of our sampling methods: nested sampling (NS) and the No U-Turn Sampler (NUTS), both with and without neural transport (NeuTra). Reported are the wall time per effective sample size (ESS), the number of gradient or likelihood (in the case of NS) evaluations per ESS, and the number of divergences normalized by the ESS. NS is run 4 times with and without NeuTra to estimate the $1\sigma$ errorbars of performance indicators, whereas with NUTS they are computed using the chain-to-chain variation.}
    \label{tab:gauss-toy-results}
\end{table}

To compare the performance of NUTS and NS, both with and without NeuTra, we use the metrics of wall time per effective sample size (ESS), and likelihood or gradient evaluation per ESS. For NUTS, as ESS can be different for each parameter, the minimum ESS is used, following~\cite{hoffman2014no}. The calculation of ESS for MCMC and NS is explained in~\cref{appendix:ESS}. We note that the ESS per evaluation metric is not directly comparable between NS and NUTS because gradient evaluations are required for NUTS and not for NS. We also include the number of divergences encountered per ESS for NUTS runs, as these indicate regions of high curvature that often cannot be reliably sampled by MCMC~\cite{Betancourt:2017ebh}.

We show the marginalized posteriors for $\mu_1$ and $\mathcal{C}_1$ in \cref{fig:toy}, produced using \texttt{corner}~\cite{corner}. We see that the $68\%$ and $95\%$ highest posterior density region contours, as well as the probability densities of these parameters, all show a high degree of overlap. Our results also agree with the ground truth. The full corner plot for all parameters is shown in \cref{appendix:corner}, and they exhibit the same behaviors.

Our numerical results are summarised in \cref{tab:gauss-toy-results}. For the evaluation per ESS, we see that NeuTra improves the performance of NS by a factor of $\usim 2$ and that of NUTS by a factor of $\usim 9$. The improvements in wall time are smaller; this is because the normalizing flow used for NeuTra has non-zero evaluation time. However, we expect the improvement in wall time and the number of likelihood function evaluations to converge for real-life problems with computationally expensive likelihood functions. Our results thus indicate that for problems with cheap likelihood functions, NeuTra--NUTS yields the best performance. 

We also find that the evidence computed using nested sampling, shown in~\cref{appendix:NS_logZ}, matches the ground truth value of $-11.14$. The mean ELBO from the last 100 epochs of neural transport training, $-11.21$, is also close to the ground truth value. This demonstrates that NeuTra can be applied to NS to accelerate Bayesian evidence computation and model comparison.

\subsection{Application to Neutrino Non-Standard Interactions}
\label{subsec:nsiexp}

\subsubsection{NSI Model}

We construct our likelihood using the NR result from XENON1T~\cite{XENON:2020gfr} and the ER measurement from PandaX-4T~\cite{Lu:2024ilt}. These measurements probe complementary parts of the NSI parameter space, allowing us to make stronger inferences compared to using either result alone.

For the XENON1T NR measurement, the data contains both solar neutrino events and background events. As our likelihood model, we use a Poisson distribution with expectation value $\lambda \equiv \lambda_{\mrm{NR}} + \lambda_{\mrm{bkg}}$. Here, $\lambda_{\text{NR}}$ is the expected number of NR events, calculated using \texttt{SNuDD} via \cref{eq:dr_gen}, and $\lambda_{\mrm{bkg}}$ is the expected number of background events. As given in~\cite{XENON:2020gfr}, we use  $\lambda_{\mrm{bkg}} = 5.38$ and take the number of observed NR events to be $N_{\mrm{NR}} = 6$.

For the PandaX-4T ER result, we use their inferred result on the number of solar neutrinos arriving at the detector per second per unit area, which is proportional to the number of ER events.
As in the NR case, we use \texttt{SNuDD} to compute the number of expected ER events for a given set of NSI parameters. We then compare this number to the one calculated in the absence of novel phenomena using the ratio between the two values, $r_\mathrm{ER}$.  We model the distribution of this ratio as a normal distribution truncated at zero at the lower end with mean and standard deviation both given by $1.72$. The truncation is to indicate that this physical quantity is strictly positive. The value $1.72$ is calculated according to the mean and uncertainty reported in \cite{Lu:2024ilt}. The above procedure gives rise to an additional likelihood term that can be used to make inferences on the number of ER events.

We physically motivate the priors on the NSI parameters below. Firstly, we expect that the intrinsic strength of the new neutrino interaction for any flavor combination, captured by $\nsi{\alpha\beta}$, should not have any preferred value. Hence, it should be uniformly distributed over a wide range, which we take to be $\left(-5,5\right)$. Secondly, given a value for this interaction strength, we do not expect any preference for the specific particle with which the neutrino interacts. Thus, we consider the prior to be uniformly distributed over the surface of the sphere in the spherical NSI space defined by the fixed radius $\nsi{\alpha\beta}$. As such, we take $\sin \eta$ to be uniformly distributed over $(-1, 1)$. The polar angle $\varphi$ is uniformly distributed in the range $(-\pi/2, \pi/2)$ instead of the usual $(0, 2 \pi)$ since the `radius' of the sphere, given by $\nsi{\alpha \beta}$, is allowed to take on negative values. The full model is summarized as follows:
\begin{equation}
\begin{split}
\hfill
   \varepsilon_{\alpha\beta} &\sim \text{Uniform}(-5, 5)\,,\\
    \varphi &\sim \text{Uniform}(-\pi/2, \pi/2)\,,\\
    \sin\eta &\sim \text{Uniform}(-1,1)\,, \hfill\\
    \hfill N_{\text{NR}} &\sim \text{Poisson}(\lambda_{\mathrm{bkg}} + \lambda_{\text{NR}}(\varepsilon_{\alpha\beta}, \eta, \varphi))\,, \\
    r_{\mathrm{ER}} &\sim \text{TruncatedNormal}(1.72, 1.72^2, 0, \infty)\,. \hfill
\end{split}
\label{eq:model}
\end{equation}

\begin{figure}[t!]
    \centering
    \includegraphics{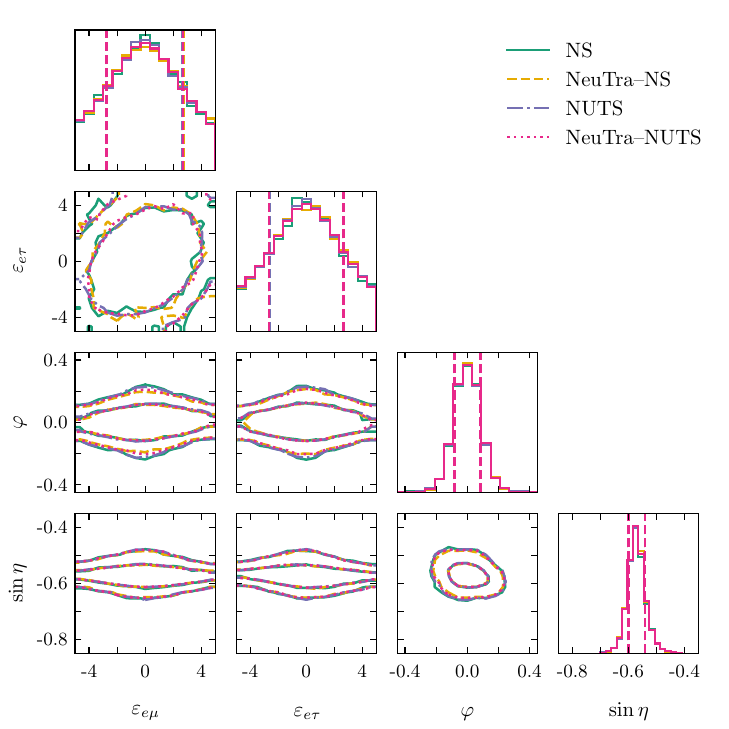}
    \caption{The $68\%$ and $95\%$ contours for the inferred posteriors for the neutrino non-standard interactions parameters. Our sampling methods are nested sampling (NS) and the No U-Turn Sampler (NUTS), both with and without neural transport (NeuTra). Two of the six $\nsi{\alpha\beta}$ are selected to be shown for visualization purposes; the full corner plot is given in \cref{appendix:corner}. As desired, all contours show a high degree of overlap. Probability densities are also shown with  $68\%$ equal-tailed intervals indicated as dashed lines. The intervals from all methods are almost identical.}
    \label{fig:nsi}
\end{figure}

We evaluate $10^4$ likelihoods with the NSI parameters randomly sampled from the priors given in \cref{eq:model}. We perform these calculations using both an Intel Xeon Gold 6230 CPU and an Nvidia Titan V GPU.
As in \cref{subsec:toyexp}, we implement NeuTra with Block Neural Autoregressive Flows, this time using three flows with hidden dimensions $[10, 10]$. The NeuTra model is trained on ELBO estimated using $30$ points for $10^4$ epochs.

For the NS runs, we use $1200$ live points without NeuTra and $800$ live points with NeuTra to produce similar ESSs of approximately $4200$. NS runs are halted when the area of the Bayesian evidence integral covered by the remaining live points drops below $2\%$; this was increased from the \texttt{ultranest} default of $1\%$ due to computational expense. We separately run NS with GPU disabled to compare the performance. This run has a smaller ESS of $\usim 800$ due to computational constraints; this causes the ESS per evaluation to be higher. 
When calculating improvement factors, we therefore multiply the wall time per ESS of the CPU NS run by $2.4$ to match the likelihood function evaluation per ESS with that of the GPU run.

For the NUTS runs, we generate $10^5$ samples both with and without NeuTra. 
We find that manually setting the step length parameter reduces the number of divergences encountered, and thus set the step size to $0.12$ and $0.06$ for NUTS with and without NeuTra, respectively. The samples are generated over $20$ chains, with $2000$ warm-up steps and $5000$ sampling steps each.

\subsubsection{Results}

\begin{table}[t!]
\renewcommand{\arraystretch}{1.5}
    \begin{tabular*}{\textwidth}{@{\extracolsep{\fill}}lccc}
    \toprule
    \textbf{Method}       & \textbf{Wall Time/ESS~[s]}  & \textbf{Eval./ESS} & \textbf{Div./ESS [$\mathbf{\times 10^{-4}}$]}\\
    \midrule
     NS (CPU, scaled)           & $130 \pm 20$ & $1500 \pm 300$  & N/A \\
    NS (GPU)          & $9 \pm 3$ & $1500 \pm 600$  & N/A \\
    NeuTra--NS (GPU)   & $0.9 \pm 0.1$ & $150 \pm 20$          & N/A \\
    \midrule[0.25pt]
    NUTS (GPU)        & $5.6 \pm 0.2$ & $195 \pm 7$ & $11 \pm 5$\\
    NeuTra--NUTS (GPU) & $2.2 \pm 0.2$ & $80 \pm 7$ & $2.3 \pm 0.9$\\
    \bottomrule
  \end{tabular*}
  \vspace{1ex}
  
  \caption{Neutrino non-standard interaction model performance of our sampling methods: nested sampling (NS) and the No U-Turn Sampler (NUTS), both with and without neural transport (NeuTra). Reported are the effective sample size (ESS) per wall time, the gradient or likelihood (in the case of NS) evaluation per ESS, and the number of divergences normalized by the ESS. The wall time and number of evaluations per ESS are scaled by a factor of 2.4 so that the evaluations per ESS are consistent between CPU and GPU runs. NS is run 6 times with and without NeuTra to estimate the $1\sigma$ errorbars of performance indicators, whereas with NUTS they are computed using the chain-to-chain variation.} 
    \label{tab:sampler-table}
\end{table}

We compare the performance of NS and NUTS, both with and without NeuTra, using the same metrics as in \cref{subsec:toyexp}. In addition, we assess the GPU acceleration by comparing the time taken for a single likelihood evaluation. 
When the GPU is enabled, we find that one such evaluation takes $3.84 \pm 0.01\,\text{ms}$ compared to $67 \pm 4\,\text{ms}$ when it is disabled. Using the GPU therefore results in a speedup of $\usim 20$ times compared to running on the CPU. 

We show the marginalized posteriors for a subset of the NSI parameters in \cref{fig:nsi}, produced using \texttt{corner}~\cite{corner}. As in \cref{subsec:toyexp}, the $68\%$ and $95\%$ highest posterior density region contours, as well as the probability densities, all agree. Due to the complexity of the model, no ground truth contour is derived. The full corner plot for all NSI parameters is shown in \cref{appendix:corner} and all demonstrate the same level of agreement.

We summarize our performances in \cref{tab:sampler-table}. For the NUTS sampler, we compute the ESS separately for each dimension. We use the minimum ESS across all dimensions, which is either $\varphi$ or $\sin\eta$, to characterize the performance following~\cite{hoffman2014no}. We find that employing NeuTra to NUTS improves the efficiency by a factor of $2$ and reduces the number of divergences by a factor of $5$. Applying NeuTra to NS accelerates it by a larger factor of 10. We therefore find that NeuTra--NS provides the greatest performance in the NSI problem.

Unlike for our synthetic problem where NeuTra--NUTS was more performant, NeuTra--NS is the better choice for our NSI problem. We believe that this is partially due to the computational expense of gradient evaluations given the complexity of our likelihood function. With information about the relative cost of gradient evaluations versus likelihood calls without automatic differentiation, it is possible to use the evaluations per effective sample information in~\cref{tab:gauss-toy-results,tab:sampler-table} to estimate which method would have a shorter runtime. It should however be noted that the ratio of evaluations per effective sample does not stay constant between the two problems, indicating that the specific structure of the posterior plays a role as well. This highlights how the optimal strategy for sampling statistical problems needs to be evaluated on a case-by-case basis.

Nevertheless, as with the synthetic problem in~\cref{subsec:toyexp}, we find the Bayesian evidence to be consistent between NS and NeuTra-NS runs. These values are shown in~\cref{appendix:NS_logZ} and are also close to the mean of the ELBO from the last 100 epochs of NeuTra training, $-10.04$.

Our result represents the first NSI parameter space scan for direct detection experiments whereby all the parameters are allowed to vary simultaneously. A similar scan was carried out in~\cite{Amaral:2023tbs}; however, only one parameter was permitted to vary at once. Our approach instead captures the full complexity of this space, allowing us to draw full multi-dimensional credible regions. In doing so, we have taken an `agnostic' EFT approach to the problem, in which we are neutral about the operators coming into play at the new physics scale, with no single one dominating. Similar studies were conducted in, for instance, Refs.~\cite{Esteban:2018ppq,Coloma:2019mbs,Coloma:2023ixt}. We present the $1$D $90\%$ credible level intervals on each of our NSI parameters in \cref{tab:nsi_intervals}, obtaining these by marginalizing over all other parameters. We use the marginalized posteriors from the result of the NeuTra-NUTS method, fully visualized in \cref{fig:nsi-full}.

\begin{table}[t!]
\renewcommand{\arraystretch}{1.5}
\begin{center}
    \begin{tabular*}{0.5\textwidth}{@{\extracolsep{\fill}}cc}
    \toprule
    \textbf{NSI Parameter} & \textbf{Credible Interval} \\
    \midrule
          $\nsi{ee}$  & $[-3.67, 3.73]$ \\
          $\nsi{e\mu}$  & $[-3.41, 3.44]$ \\
          $\nsi{e\tau}$  & $[-3.33, 3.40]$ \\
          $\nsi{\mu\mu}$  & $[-3.86, 3.83]$ \\
          $\nsi{\mu\tau}$  & $[-3.70, 3.75]$ \\
          $\nsi{\tau\tau}$  & $[-3.84, 3.74]$ \\
          $\eta$  & $[-37.6^\circ,\,-32.0^\circ]$ \\
          $\varphi$  & $[-5.88^\circ,\,6.15^\circ]$ \\
    \bottomrule 
  \end{tabular*}
\end{center}
    \caption{One-dimensional $90\%$ credible level intervals on each of our NSI parameters. The interval on any one parameter is computed following marginalization over all other parameters.}
    \label{tab:nsi_intervals}
\end{table}

When taking our agnostic EFT approach, we find the expected result that the overall strength of neutrino NSI, characterized by the parameters $\nsi{\alpha\beta}$, is poorly constrained. This is due to the large number of interferences that can occur between these parameters, leading to signal cancellations. Nevertheless, in this scheme, we are able to tightly constrain the `direction' of NSI, finding narrow intervals for both $\eta$ and $\varphi$. 

This latter result indicates that, within this agnostic scheme, neutrino NSI cannot take place purely with the electron ($\varphi = \pm 90^\circ)$, nor can they take place only with up- or down-quarks (characterized at $\varphi = 0^\circ$ by the angles $\eta \approx 26.6^\circ$ and $\eta \approx 63.4^\circ$, respectively). Instead, we are constrained to lie tightly along the proton direction in the charged plane and also along where there is a cancellation between proton and neutron NSI in the neutrino-nucleus cross section for a xenon target, occurring at $\eta \approx -35^\circ$ (see Ref.~\cite{Amaral:2023tbs} for details). A similar result was found in Ref.~\cite{Coloma:2023ixt}, wherein combining data from processes sensitive to nuclear and electronic processes---in their case through results respectively from spallation source and solar neutrino/reactor experiments---led to tight constraints on the direction of NSI centred around our intervals.

However, a key difference between our results and those of Ref.~\cite{Coloma:2023ixt} is that we are able to achieve these intervals using only a \textit{single type} of experiment. Direct detection experiments are uniquely placed to probe both nuclear and electronic signals, allowing us to simultaneously test neutrino NSI with nucleons and electrons. While current experiments are unable to single-handedly provide leading constraints in this the NSI space, we expect that future experimental data will be an invaluable addition to global analyses~\cite{Amaral:2023tbs}. In this case, having robust methods to include direct detection experiments in these studies promptly, such as those we provide in this work, will be welcomed by the community.

\section{Conclusions}
\label{sec:conc}

We have achieved significant performance improvements in Bayesian inference compared to techniques traditional to the astroparticle physics community. We accomplished this by leveraging GPU acceleration, automatic differentiation, and neural-network-guided reparameterization, benchmarking their performances against nested sampling alone.
We have made inferences in the multi-dimensional parameter space of neutrino non-standard interactions using the astroparticle experiments XENON1T and PandaX-4T. We have achieved a factor $\usim 100$ performance boost compared to nested sampling alone without compromising the accuracy of the results, also producing the first parameter space scan for direct detection experiments where all NSI parameters are allowed to vary simultaneously.

We began by using a multivariate Gaussian model to evaluate the performances of nested sampling and the No U-Turn Sampler (NUTS), both with and without neural transport. We found that neural transport improved their performances by factors of $\usim 2$ and $\usim 9$, respectively. For this synthetic problem, NUTS was more performant at posterior distribution sampling than nested sampling.

Applying these methods to the NSI parameter space, we found that GPUs accelerated likelihood evaluations by a factor of $\usim 20$ and that NUTS was faster than nested sampling by a factor of $\usim 2$. Finally, reparameterization using neural transport, whereby our posteriors were first mapped to Gaussian geometries, gave us a total factor of $\usim 60$ improvement with NUTS and $\usim 100$ with nested sampling compared to nested sampling on the CPU without neural transport. Unlike in the Gaussian fit, the performance of NUTS was lower than that of nested sampling when neural transport was employed, demonstrating that the optimal method depends on the specific problem at hand.
In addition, since nested sampling allows for the computation of the Bayesian evidence, this work represents a way to accelerate model comparison while retaining compatibility with existing nested sampling implementations that are widely used in the natural sciences. 

Within the NSI space, this result is the first scan for direct detection experiments wherein all parameters are allowed to vary at once. We took an agnostic effective field theory approach in which not only the strength of potential new interactions with neutrinos can change but also the specific particles they can interact with can vary. We found that any individual parameter controlling the interaction strength is only loosely constrained with direct detection data, but that the particles with which new interactions can take place are well constrained at a level comparable to that achieved using dedicated neutrino experiments.

Our results underscore the potential of advanced computational techniques to transform inference and model comparison in astroparticle physics. The performance improvements we have achieved will aid in diagnosing convergence issues, incorporating additional experimental data, and producing timely studies. Moreover, these methods can be extended to other areas in astroparticle physics featuring multi-dimensional parameter spaces.

\subsection{Limitations and Future Work}
\label{subsec:lim_future}

For our neutrino physics experiments, the computational resource used to train the neural transport model was insignificant compared to the sampling process. This may not be true if only a small posterior sample is needed, and thus whether neural transport would yield performance improvements would depend on the desired posterior sample size. We note that the effectiveness of neural transport is dependent on the ability of the trained normalizing flow to represent the desired posterior distribution~\cite{pmlr-v202-grenioux23a} and that Neural transport requires manual tuning for individual problems. For future work, we will assess how comparing the Bayesian evidence evaluated using nested sampling and the ELBO from neural transport training can be used as a diagnostic for both the convergence of nested sampling and the training of the normalizing flow. In this work, while each likelihood evaluation is internally vectorized, we have also not worked on vectorizing our likelihood functions for batched evaluation. Vectorized likelihood functions might lead to GPUs having an even greater speedup over CPUs for nested sampling, and we hope to address this in the future.

Finally, we are applying the methods in this work to so-called \textit{global fits} of neutrino properties. In these fits,  multi-dimensional parameter spaces are independently scanned for multiple experiments, and they are produced by collaborations such as  \texttt{NuFIT} \cite{Esteban:2020cvm}.
Combining the constraining power from many experiments---especially those with multiple measurement channels such as DUNE~\cite{DUNE:2020ypp}, PandaX~\cite{PandaX:2014mem}, and XENONnT~\cite{XENON:2024wpa}---will make our methods particularly pertinent. This is because each experiment will have an associated complex posterior geometry to traverse that can benefit from the techniques we have explored.

\section{Data Availability Statement}

No new data was created or analyzed in this study. We provide our code in \href{https://github.com/RiceAstroparticleLab/ML-NSI}{\faGithub}.

\begin{ack}
We thank Aar\'on Higuera for insightful discussions throughout this work. We also thank Fangda Gu, Ray Hagimoto, Aar\'on Higuera, Ivy Li, and Andre Scaffidi for their comments on the manuscript. This work is supported by the Department of Energy AI4HEP program, Rice University, and the National Science Foundation CAREER award PHY-204659. We thank Nvidia for supplying us with the Titan V GPUs used in this work.
\end{ack}

\section*{References}
\bibliographystyle{unsrt}
\bibliography{biblio}

\newpage

\appendix

\section{Reparameterizing Probability Density Functions}
\label{appendix:reparam}

Let $\bmath{f}: \mathbb{R}^n \rightarrow \mathbb{R}^n$ denote an arbitrary coordinate transformation onto the same space. This function takes the vector of random variables $\bmath{X} \equiv (X_1, X_2, \dots, X_n)^\intercal$ and maps it onto the transformed random variable vector $\bmath{Y} \equiv \bmath{f}(\bmath{X}) \equiv (Y_1, Y_2, \dots, Y_n)^\intercal$. The change in the volume element from the first basis to the second is characterized by the determinant of the Jacobian
\begin{equation}
    \label{eq:jac}
    \frac{\partial\bmath{f}}{\partial\bmath{x}} \equiv \begin{pmatrix}
    \frac{\partial f_1}{\partial x_1} & \cdots & \frac{\partial f_1}{\partial x_n} \\
    \vdots & \ddots & \vdots \\
    \frac{\partial f_n}{\partial x_1} & \cdots & \frac{\partial f_n}{\partial x_n}
    \end{pmatrix}\,.
\end{equation}

Consider the joint-probability density of the transformed and original variables, respectively $\rho_Y(\bmath{y})$ and $\rho_X(\bmath{x})$. Since probabilities within equal volumes must remain fixed, we have that
\begin{equation}
    \int_{\bmath{f}(\bmath{\Omega})} \rho_{\bmath{Y}}(\bmath{y})\,\mathrm{d}^n\boldsymbol{y} = \int_{\bmath{\Omega}} \rho_{\bmath{Y}}[\bmath{f}(\bmath{x})] \left|\frac{\partial\bmath{f}}{\partial\bmath{x}}\right|\mathrm{d}^n\boldsymbol{x}\,,
\end{equation}
where $\bmath{\Omega}$ is the integration domain with respect to the original variables. We can thus identify the probability density
\begin{equation}
    \rho_{\bmath{X}}(\bmath{x}) = \rho_{\bmath{Y}}[\bmath{f}(\bmath{x})] \left|\frac{\partial\bmath{f}}{\partial\bmath{x}}\right|\,,
\end{equation}
which is the form of \cref{eq:latent-transform}. Note that a similar procedure can be used to relate $\rho_{\bmath{Y}}(\bmath{y})$ to $\rho_{\bmath{X}}(\bmath{x})$ via
\begin{equation}
    \rho_{\bmath{Y}}(\bmath{y}) = \rho_{\bmath{X}}[\bmath{f}^{-1}(\bmath{y})] \left|\frac{\partial\bmath{x}}{\partial\bmath{f}}\right|\,,
\end{equation}
where $\left|\partial\bmath{x}/\partial\bmath{f}\right|$ is the determinant of the inverse of the Jacobian given in \cref{eq:jac}.

\section{Ground Truth Bayesian Evidence and Contour for Synthetic Model}
\label{appendix:bayesian}

We consider a generalized version of the synthetic model we introduced in \cref{subsec:toyexp}, with parameters $\bmath{\theta} \in \mathbb{R}^k$ and data matrix $\bmath{X} \in \mathbb{R}^{k \times n}$, such that $\bmath{X} \equiv (\bmath{x}_1, \bmath{x}_2, \dots, \bmath{x}_n)^\intercal$ with $\bmath{x}_i \in \mathbb{R}^k$. Each random vector $\bmath{x}_i$ represents a vector of random variables sampled from a $k$-dimensional Gaussian. The likelihood is then given by
\begin{equation}
    \ell(\bmath{\theta}; \bmath{X}) = \prod_{i=1}^{k}\prod_{j=1}^{n} \frac{1}{\sqrt{2\pi \sigma_i^2}} \exp\left[-\frac{1}{2}\left(\frac{X_{ij} - \mu_i}{\sigma_i^2}\right)^2\right]\,.
\end{equation}
As $\sigma_i^2$ needs to be strictly positive, we instead define priors on $\mathcal{C}_i = \log \sigma_i^2$. Our parameters are then $\bmath{\theta} \equiv (\mu_1, \dots, \mu_{k/2}, \mathcal{C}_1, \dots, \mathcal{C}_{k/2})^\intercal$. We use the same model as \cref{eq:toy_model}, such that our prior distribution is given by
\begin{equation}
    \pi(\bmath{\theta}) = \prod_{i=1}^{k} \frac{1}{\sqrt{2 \pi \tilde{\sigma}^2}} \exp\left[-\frac{\mu_i^2}{2 \tilde{\sigma}^2}\right]\exp\left[-\frac{\mathcal{C}_i^2}{2 \tilde{\sigma}^2}\right]\,,
\end{equation}
where the variance for all parameters is set to $\tilde{\sigma}^2 = 10$.

We first wish to calculate the Bayesian evidence integral,  given by
\begin{equation}
    \mathcal{Z} \equiv \int \ell(\bmath{\theta}; \bmath{X}) \funop{\pi(\bmath{\theta})} \mathrm{d}^k \bmath{\theta}\,.
    \label{eq:bayesian}
\end{equation}
To reduce computational cost, we perform the integrals over $\bmath{\mu}$ analytically. This halves the number of integrals that need to be done numerically and improves the performance scaling from $O(N^{2k})$ to $O(N^k)$, given $N$ integration sample points per dimension. 

We may write \cref{eq:bayesian} as
\begin{equation}
    \mathcal{Z} = \frac{1}{(2\pi)^{n k / 2}}\frac{1}{(2\pi \tilde{\sigma}^{2})^{k}} \prod_{i=1}^{k} \iint e^{-n \mathcal{C}_i} e^{-\mathcal{E}_{i}(\mu_i)}\,\mathrm{d}^{k/2} \bmath{\mu}\,\mathrm{d}^{k/2}\bmath{\mathcal{C}}\,,
\end{equation}
where the exponent $\mathcal{E}_i$ has been defined to be
\begin{equation}
    \mathcal{E}_i(\mu_i) \equiv a_i \mu_i^2 + b_i \mu_i + c_i\,,
\end{equation}
with
\begin{equation}
    a_i \equiv \frac{1}{2}\left(\frac{1}{\tilde{\sigma}^2} + n e^{-2 \mathcal{C}_i}\right)\,, \quad b_i \equiv - e^{-2 \mathcal{C}_i}\sum_{j=1}^{n} y_{ij}\,, \quad c_i \equiv \frac{1}{2}\left(\frac{\mathcal{C}_i^2}{\tilde{\sigma}^2} + e^{-2 \mathcal{C}_i} \sum_{j=1}^{n} y_{ij}^2\right)\,.
\end{equation}
Integrating over any one particular $\mu_i$ then gives us
\begin{equation}
    \mathcal{I}_i \equiv \int_{-\infty}^{\infty} e^{-\mathcal{E}_i(\mu_i)}\,\mathrm{d} \mu_i =
     \sqrt{\frac{\pi}{a_i}}\exp\left(\frac{b_i^2 - 4a_ic_i}{4a_i}\right)\,,
\end{equation}
such that 
\begin{equation}
    \mathcal{Z} = \frac{1}{(2\pi)^{n k / 2}}\frac{1}{(2\pi \tilde{\sigma}^{2})^{k}} \prod_{i=1}^{k} \int e^{-n \mathcal{C}_i} \mathcal{I}_{i} \,\mathrm{d}^{k/2}\bmath{\mathcal{C}}\,.
\end{equation}
The remaining $k/2$ integrals on $\mathcal{C}_i$ can be done numerically.

\begin{figure}
    \centering
    \includegraphics{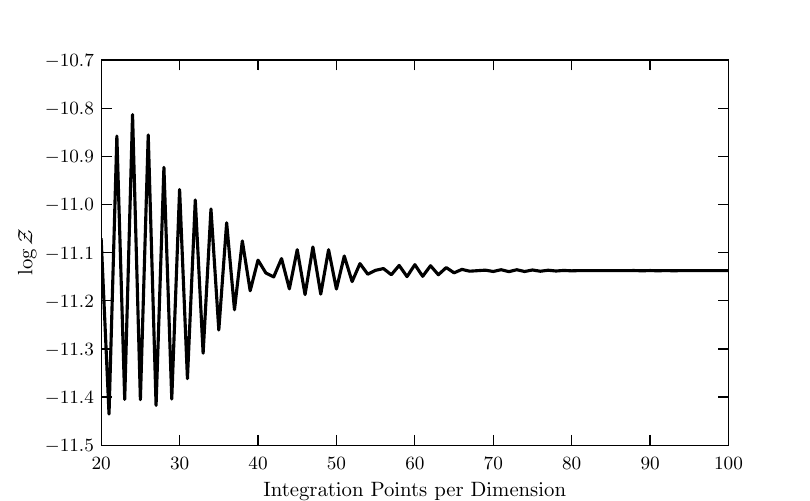}
    \caption{Log of Bayesian evidence integral $\log \mathcal{Z}$ for synthetic model, \cref{eq:bayesian}, with number of integration points per parameter dimension. For this model, the number of dimensions is $k = 3$ and the number of data points is $j = 2$. The integral converges to the quoted value of $\log \mathcal{Z} \approx -11.14$ after approximately $60$ points.}
    \label{fig:logZ}
\end{figure}

For the numerical integration, we use the \texttt{torchquad} package~\cite{Gomez_torchquad_Numerical_Integration_2021}. In the particular synthetic model described in \cref{subsec:toyexp}, we have $k = 3$ dimensions and $j=2$ data points. We use the trapezoid procedure with a domain of $[-50, 50]$ for both $\mu_i$ and $\mathcal{C}_i$, and we use $151$ integration points per dimension. The domain was chosen to be far from the contours; this was verified by increasing the integration domain to $[80, 80]$, scaling the integration points per dimension to $241$, and checking that the value of the integral does not appreciably change. We ensure the integral has converged by verifying that $\log \mathcal{Z}$ does not change when the number of integration points per dimension is increased, as shown in~\cref{fig:logZ}. We converge to the quoted value of $\log \mathcal{Z} \approx -11.14$ after approximately $60$ points, and subsequently conservatively choose $151$ points for numerical integrals used in this paper. This corresponds to $151^3$ points for the  $\log \mathcal{Z}$ integral, and $151^2$ points for the $\mu_1$ versus $\mathcal{C}_1$ contour. 

To compute the ground truth contour, the two variables on which the contour is projected are excluded from the integral in~\cref{eq:bayesian}, producing the quantity
\begin{equation}
    \mathcal{F}(\theta_u, \theta_v) \equiv \int \ell(\theta_u, \theta_v, \bmath{\theta'}; \bmath{X}) \funop{\pi(\theta_u, \theta_v, \bmath{\theta'})} \,\mathrm{d}^{k-2} \bmath{\theta'}\,,
\end{equation}
where $\theta_u$ and $\theta_v$ are the variables that are used to produce the contour, and $\bmath{\theta'}$ is the set of variables excluding $\theta_u$ and $\theta_v$. $\mathcal{F}(\theta_u, \theta_v)$ is then computed using the same procedure described above for each set of $\theta_u$ and $\theta_v$. A $\rho\%$ contour level can be described by the equation $\varsigma_{\rho\%} = \mathcal{F}(\theta_u, \theta_v)$, where
\begin{equation}
    \frac{\int \mathcal{F}'(\theta_u, \theta_v) \,\mathrm{d} \theta_u \,\mathrm{d} \theta_v}{\int \mathcal{F}(\theta_u, \theta_v) \,\mathrm{d} \theta_u \,\mathrm{d} \theta_v} = \rho\%\,, \quad \text{with} \quad \mathcal{F}'(\theta_u, \theta_v) \equiv
    \begin{cases}
        \mathcal{F}(\theta_u, \theta_v) & \text{if} \quad \mathcal{F}(\theta_u, \theta_v) > \varsigma_{p\%}\\
        0 & \text{if} \quad \mathcal{F}(\theta_u, \theta_v) \leq \varsigma_{p\%}
    \end{cases}\,.
\end{equation}
We show the ground truth $\mu_1$--$\mathcal{C}_1$ contour in \cref{fig:toy}.

\section{Effective Sample Size Calculation}
\label{appendix:ESS}

The effective sample size (ESS) for MCMC chains, as implemented in \texttt{ArviZ}, is given by~\cite{Vehtari2019RankNormalizationFA, arviz_2019}
\begin{equation}
    N_{\mathrm{eff}} = \frac{MN}{\tau}\,, \quad
    \tau = -1 + 2 \sum^{K}_{t=0} P_{t'}\,, \quad \text{and} \quad
    P_{t'} = \rho_{2t'} + \rho_{2t' + 1}\,,
\end{equation}
where $M$ is the number of chains, $N$ is the number of samples per chain, $\rho_{t}$ is the estimated autocorrelation at time $t$, and $K$ is the largest integer for which $P_{K} = \rho_{2K} + \rho_{2K+1}$ remains positive. As nested sampling produces weighted samples, the effective sample size is instead computed using Kish's design effect~\cite{alma991010178329705251}, as implemented in \texttt{ultranest}~\cite{2021JOSS....6.3001B}:
\begin{equation}
    D_{\mathrm{eff}} = 1 + \frac{1}{N}\sum^N_{i=1}(N w_i - 1)^2 \qquad \mathrm{and} \qquad 
    N_{\mathrm{eff}} = \frac{N}{D_{\mathrm{eff}}}\,,
\end{equation}
where $w_i$ are normalized weights that sum to 1 and $N$ is the total number of weighted samples.

\section{Full Corner Plots}
\label{appendix:corner}

\begin{figure}[h!]
    \centering
\includegraphics{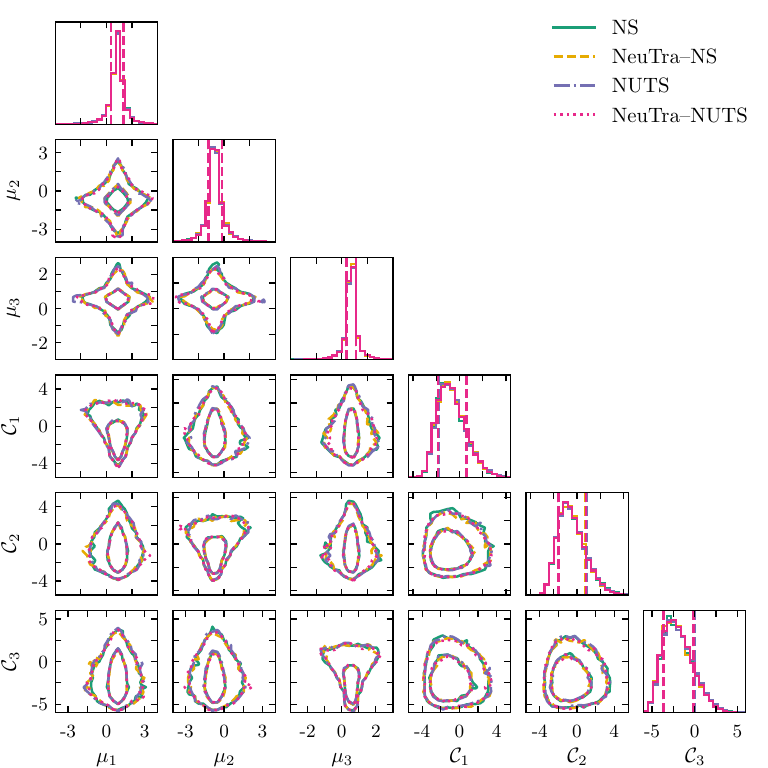}
    \caption{The same as in \cref{fig:toy} but for the full synthetic model parameter space.}
    \label{fig:toy-full}
\end{figure}

\begin{figure}[h!]
    \centering
    \includegraphics{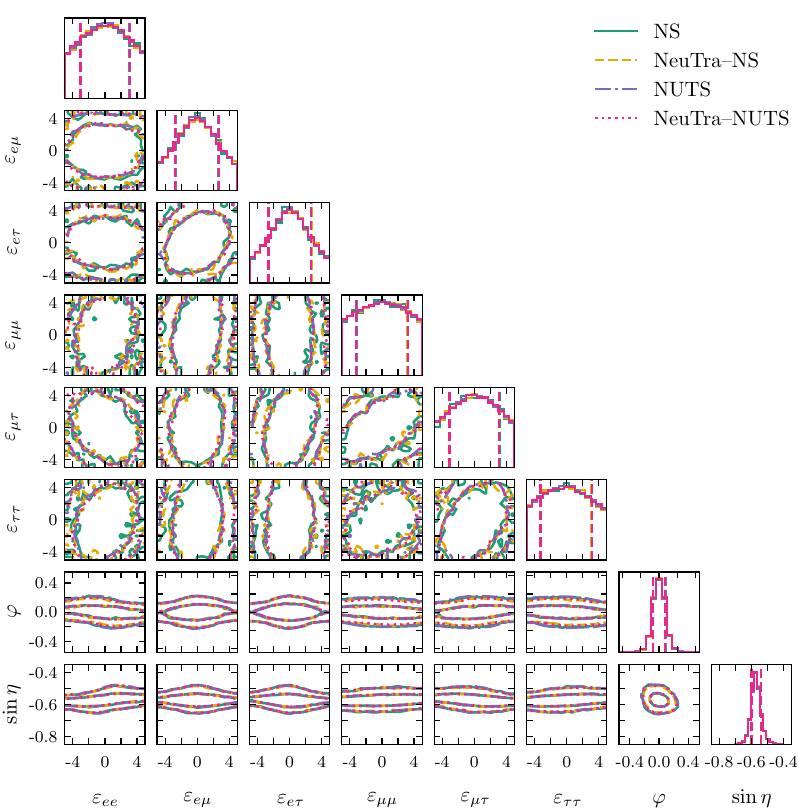}
    \caption{The same as in \cref{fig:nsi} but for the full NSI parameter space.}
    \label{fig:nsi-full}
\end{figure}

\clearpage

\section{Bayesian Evidence from Nested Sampling Runs}
\label{appendix:NS_logZ}

The Bayesian evidences from the Gaussian fit with unknown mean and variance and the neutrino physics model are shown in~\cref{tab:evidence_integral} and~\cref{tab:evidence_integral_nsi}, respectively.

\begin{table}[h!]
\renewcommand{\arraystretch}{1.5}
    \begin{tabular*}{\textwidth}{@{\extracolsep{\fill}}cccc}
    \toprule
    \textbf{Run number}       & \textbf{$\boldsymbol{\log\mathcal{Z}}$ (NS)}  & \textbf{$\boldsymbol{\log\mathcal{Z}}$ (NeuTra--NS)}\\
    \midrule
    0 & $-11.17 \pm 0.07$ & $-11.19 \pm 0.1$\\
    1 & $-11.1 \pm 0.1$ & $-11.13 \pm 0.06$\\
    2 & $-11.14 \pm 0.08$ & $-11.14 \pm 0.07$\\
    3 & $-11.1 \pm 0.2$ & $-11.16 \pm 0.06$\\
    \bottomrule
  \end{tabular*}
  \vspace{1ex}
  
  \caption{Log of Bayesian evidence integral for the Gaussian fit, $\log\mathcal{Z}$, computed using nested sampling (NS), with and without neural transport (NeuTra). We see that the evidence typically has a lower uncertainty in the NeuTra--NS runs. The values from all runs match both the ELBO from neural transport training, $-11.21$, and the ground truth value, $-11.14$.}
    \label{tab:evidence_integral}
\end{table}

\begin{table}[h!]
\renewcommand{\arraystretch}{1.5}
    \begin{tabular*}{\textwidth}{@{\extracolsep{\fill}}cccc}
    \toprule
    \textbf{Run number}       & \textbf{$\boldsymbol{\log\mathcal{Z}}$ (NS)}  & \textbf{$\boldsymbol{\log\mathcal{Z}}$ (NeuTra--NS)}\\
    \midrule
    0 & $-9.7 \pm 0.1$ & $-9.8 \pm 0.1$\\
    1 & $-9.89 \pm 0.08$ & $-9.8 \pm 0.2$\\
    2 & $-9.8 \pm 0.1$ & $-9.8 \pm 0.2$\\
    3 & $-9.97 \pm 0.08$ & $-9.6 \pm 0.2$\\
    4 & $-9.8 \pm 0.1$ & $-9.8 \pm 0.1$\\
    5 & $-9.7 \pm 0.08$ & $-9.9 \pm 0.1$\\
    \bottomrule
  \end{tabular*}
  \vspace{1ex}
  
  \caption{Log of Bayesian evidence integral for the neutrino physics problem, $\log\mathcal{Z}$, computed using nested sampling (NS), with and without neural transport (NeuTra). We see that the evidence typically has a lower uncertainty in the NeuTra--NS runs. The values from all runs match the ELBO from neural transport training, $-10.04$.}
    \label{tab:evidence_integral_nsi}
\end{table}

\end{document}